\documentclass[letterpaper, 10 pt, journal, twoside]{IEEEtran}

\usepackage{comment}
\usepackage[pdftex]{graphicx}
\usepackage{subfigure}
\usepackage{amsmath}
\usepackage{bbm}
\usepackage{dsfont}
\usepackage{cite}
\usepackage{color}
\usepackage{url}
\usepackage{comment}

\begin{document}

\title{SLAMER: Simultaneous Localization and Map-Assisted Environment Recognition}

\author{Naoki~Akai$^{1}$
\thanks{$^{1}$N. Akai is with the Graduate School of Engineering, Nagoya University, Nagoya 464-8603, Japan {\tt\footnotesize akai@nagoya-u.jp}}
\thanks{Digital Object Identifier (DOI): see top of this page.}
}

\markboth{IEEE Robotics and Automation Letters,~Vol.~xx, No.~x, July~202x}%
{Akai \MakeLowercase{\textit{et al.}}: SLAMER}

\maketitle

\begin{abstract}
This paper presents a simultaneous localization and map-assisted environment recognition (SLAMER) method. Mobile robots usually have an environment map and environment information can be assigned to the map. Important information for mobile robots such as no entry zone can be predicted if localization has succeeded since relative pose of them can be known. However, this prediction is failed when localization does not work. Uncertainty of pose estimate must be considered for robustly using the map information. In addition, robots have external sensors and environment information can be recognized using the sensors. This on-line recognition of course contains uncertainty; however, it has to be fused with the map information for robust environment recognition since the map also contains uncertainty owing to over time. SLAMER can simultaneously cope with these uncertainties and achieves accurate localization and environment recognition. In this paper, we demonstrate LiDAR-based implementation of SLAMER in two cases. In the first case, we use the SemanticKITTI dataset and show that SLAMER achieves accurate estimate more than traditional methods. In the second case, we use an indoor mobile robot and show that unmeasurable environmental objects such as open doors and no entry lines can be recognized.
\end{abstract}


\IEEEpeerreviewmaketitle

\section{Introduction}
\label{sec:introduction}

\IEEEPARstart{A}{utonomous} mobile robots usually have an environment map and any information can be assigned to the map.
The assigned information can be predicted if localization has succeeded because the relative position of them can be exactly known.
This map-based prediction enables robots to effectively recognize important information for autonomous navigation such as no entry zone~\cite{AkaiJRM2015}.
However, this prediction is failed when localization does not work and it might occur fail of autonomous navigation.
Uncertainty of localization must be considered to perform robust map-based environment recognition.
In addition, mobile robots have external sensors such as cameras and LiDARs and these can be used for environment recognition; however, sensor-based environment recognition also contains uncertainty.
Hence, this sensor-based recognition has to be fused with the map-based recognition for robust environment recognition.
However, mapping uncertainty also has to be considered because environment usually changes over time.
This paper presents a simultaneous localization and map-assisted environment recognition (SLAMER) method to realize simultaneous consideration of these uncertainties.

\begin{figure}[!t]
    \begin{center}
        \includegraphics[width = 85 mm]{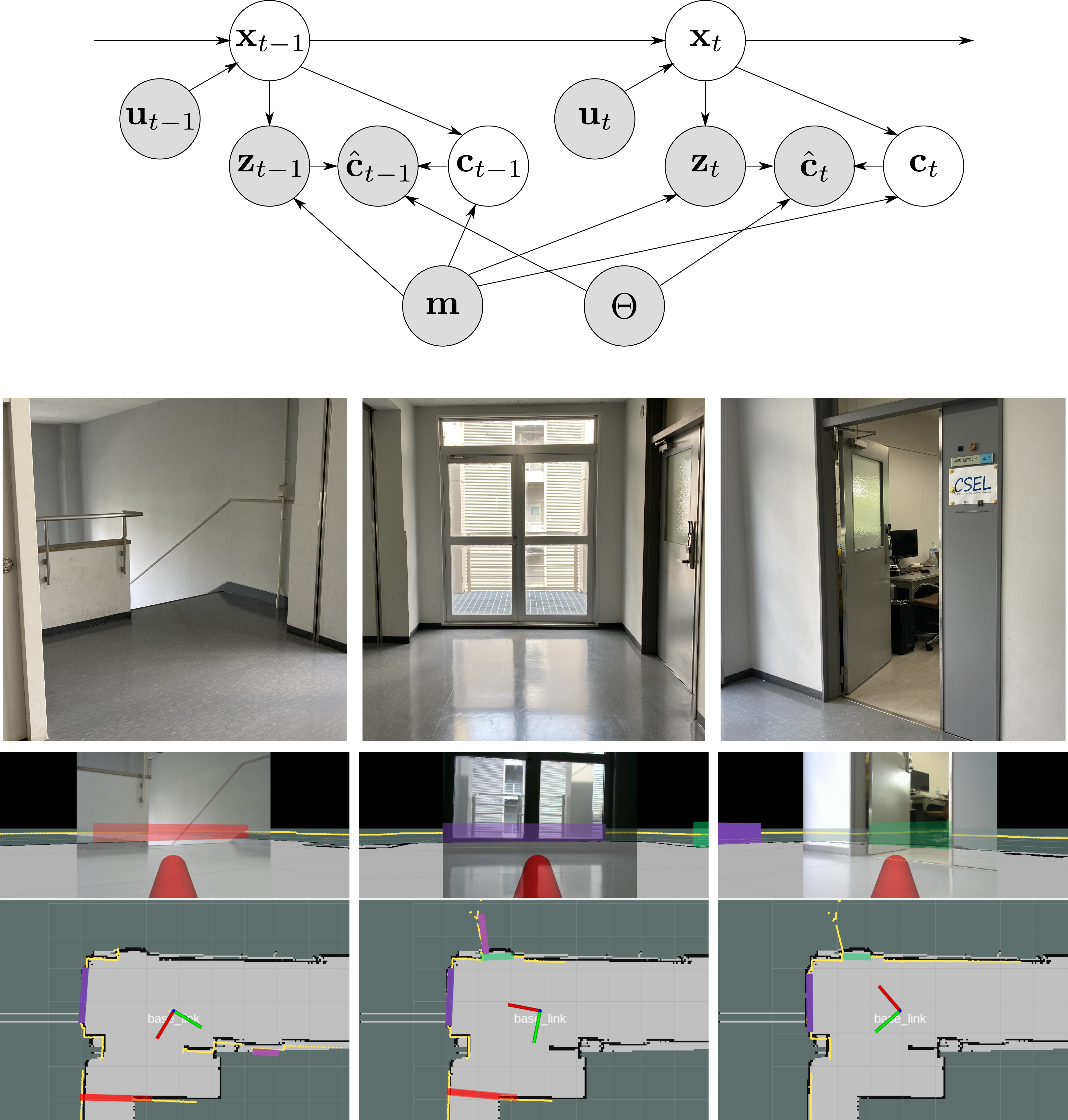}
        \caption{Top: The graphical model of SLAMER. Bottom: Estimate examples by 2D LiDAR-based SLAMER. No entry line (red), glass door (purple), and open door (green) are recognized utilizing the map assist.}
        \label{fig:concept}
    \end{center}
\end{figure}

The graphical model of SLAMER is illustrated in the top of Fig.~\ref{fig:concept}.
The white and gray nodes represent hidden and observable variables and the arrows represent dependencies where the tip variables depend on the root variables.
A pose of a robot, ${\bf x}$, and true environmental object classes, ${\bf c}$, are treated as the hidden variables, and control input, ${\bf u}$, sensor measurement, ${\bf z}$, semantic map, ${\bf m}$, environmental object recognition results, $\hat{{\bf c}}$, and hyperparameters of the recognition method, $\Theta$, are treated as the observable variables.
It should be noted that the true classes and recognition results are different.
These variables are detailed in Section~\ref{subsec:target_problem_and_variable_definition}.

In the general localization model presented in~\cite{Thrun:2005:PR:1121596}, $\hat{\bf c}$ and ${\bf c}$ do not exist.
In the SLAMER model, these variables are introduced as the observable and hidden variables, and we assume that the recognition results depend on the true classes.
This relationship enables us to estimate the true classes using the recognition results.
In addition, we assume that the true classes depend on the pose and the semantic map.
This relationship enables us to use the pose and map information to predict the true classes.
Owing to these relationships, the true object classes can be estimated while respecting to both map-based prediction and sensor-based recognition and these results can be fused with the Bayes theorem.
Consequently, uncertainties in localization, environment recognition, and mapping can be simultaneously considered.
The bottom figures of Fig.~\ref{fig:concept} shows estimate examples by 2D-LiDAR-based SLAMER.
SLAMER can recognize objects which do not provide better influence to localization such as no entry line (red), glass door (purple), and open door (green) because the SLAMER model enables to treat such objects.

In this paper, we demonstrate LiDAR-based implementation of SLAMER with two cases; urban driving car equipped with 3D LiDAR and indoor mobile robot equipped with 2D LiDAR.
In the urban driving case, we use the SemanticKITTI dataset~\cite{behley2019iccv} and show quantitative results including comparison with other methods.
The comparison show that SLAMER outperforms traditional methods; however, it cannot outperform the class prediction model (CPM) that is presented in~\cite{AkaiRA-L2020} for performing localization with semantic information in terms of localization accuracy.
This paper also discusses regarding advantage and disadvantage of SLAMER while comparing with CPM.
In this discussion, the modeling concept of both SLAMER and CPM is considered.
In the indoor mobile robot case, we use our own experimental platform to show qualitative results as shown in the bottom of Fig.~\ref{fig:concept}.
Through the demonstrations, we show that SLAMER improves both localization and environment recognition accuracy.
The contribution of this paper is twofold.
\begin{itemize}
    \item Proposing the general framework for performing simultaneous localization and map-assisted environment recognition
    \item Discussing advantage and disadvantage of SLAMER while considering relationship with CPM
\end{itemize}

The rest of this paper is organized as follow.
Section~\ref{sec:related_work} summarizes related works.
Section~\ref{sec:proposed_method} details the proposed method.
Section~\ref{sec:implementation} describes implementation ways of the proposed method.
Sections~\ref{sec:dataset_experiments} and \ref{sec:indoor_experiments} provide experimental results and discussion.
Section~\ref{sec:conclusion} concludes this work.

\section{Related work}
\label{sec:related_work}

Recent evolution of machine learning, in particular deep leaning, enables us to obtain accurate semantic information from camera images~\cite{badrinarayanan2015segnet} and/or 3D point clouds~\cite{RangeNet}.
Owing to the breakthroughs, the use of semantics in localization and SLAM becomes popular in recent.

A semantic-aware visual localization method is presented in~\cite{Stenborg2018LongTermVL}.
In~\cite{Stenborg2018LongTermVL}, the semantics-assisted measurement model is presented.
A deep-learning-based visual localization method with semantics presented in~\cite{semantic_visual_localization}.
The network used in the method takes incomplete semantic observations and predicts completed semantic subvolumes.
The method also uses joint geometric and semantic matching that confirms whether voxels are matched well in term of both geometric and semantics.
A semantic-aware visual SLAM method is presented in~\cite{SmanticVisualSLAM, QuadricSLAM}.
In~\cite{SmanticVisualSLAM}, dynamic and static keypoints can be separated with the semantics.
In~\cite{QuadricSLAM}, the geometric error term that enables to use object detection systems such as Faster R-CNN~\cite{FasterR-CNN} as sensors in SLAM is presented.
A semantic-aware localization method using a LiDAR is presented in~\cite{zaganidis2017registration, 8387438}.
In these works, the semantics is utilized in the cost function for point cloud registration to mitigate influence of outliers.
A semantic-aware SLAM method using a LiDAR is presented in~\cite{SuMa++} and it also uses the semantics in the cost function for registration.
These methods use the semantics for improving localization and mapping accuracy; however, uncertainty in semantics recognition is not considered because these uses are similar to robust estimation such as M-estimation~\cite{Hartley:2003:MVG:861369}.
In addition, map-assisted environment recognition is not focused on.

Parkison {\it et al}.~\cite{Parkison2018SemanticIC} presented the semantic ICP scan matching method with the Expectation-Maximization (EM) algorithm.
Through the EM algorithm, ambiguity of the semantics can be coped with.
The similar approaches that uses Markov random fields or conditional random fields to cope with outliers in scan matching can be seen at~\cite{DBLP:journals/corr/abs-1711-05864, DBLP:conf/rss/RamosFD07}.
Bowman {\it et al}.~\cite{7989203} presented the probabilistic data association method for semantic SLAM using the $k$-best assignment enumeration.
The $k$-best assignment enables to compute marginal assignment probabilities for each measurement landmark pair.
Atanasov {\it et al}.~\cite{doi:10.1177/0278364915596589} the localization method using the semantic observations.
They proposed the sensor model encodes the semantics via random finite sets and it realizes a unified treatment of miss detection, false alarms, and data association.
They also proposed the efficient likelihood calculation method using the matrix permanent.
These methods consider uncertainty in semantics recognition; however, our proposal extends these approaches to a simultaneous environment recognition approach which utilizes the semantic map.
Namely, semantics recognition results are improved while performing localization.

Our proposal performs simultaneous localization and environmental recognition.
A similar approach that simultaneously updates an environment map while performing localization in dynamic environments is presented in~\cite{doi:10.1177/0278364913502830}.
This probabilistic simultaneous estimation is typically implemented based on Rao-Blackwellized particle filter, e.g., FastSLAM 2.0~\cite{FastSLAM2.0}, and this implementation requires large memory cost since each particle has to have a map.
However, the proposed method does not require large memory cost since the particles do not have a map.
Wang {\it et al}.~\cite{wang_ijrr2007:_slam_mot} presented the simultaneous localization, mapping, and moving object tracking method.
This simultaneous estimation is also similar to our proposal; however, our focus is different from their focus.
The methods presented in~\cite{doi:10.1177/0278364913502830, wang_ijrr2007:_slam_mot} can be used for map update; however, SLAMER cannot be used for its purpose.
Instead, SLAMER can perform robust localization owing to simultaneous map-assisted environment recognition.

The geometric map-assisted localization method for monocular visual odometry is presented in~\cite{7814249}.
This achieves coping with uncertainties in drift and scale in the process of monocular visual odometry.
Geometric map-assisted shape matching is implemented as the measurement model.
Using such map assist in the measurement model is quite similar to our proposal; however, our target is using the semantics and it requires a framework to handle discrete variables.
An object recognition method using deep learning in a 2D occupancy grid map is presented in~\cite{8569433}.
This network could utilize map data for recognition; however, our proposal provides explicit formulation of the map-assisted environment recognition.

We proposed similar approach to SLAMER in~\cite{AkaiIROS2018}.
In~\cite{AkaiIROS2018}, two measurement classes, mapped and unmapped obstacles, are considered and these are simultaneously recognized while performing localization.
In the model, the measurement classes are introduced as the hidden variable.
This introduction is the same to that of the SLAMER model.
However, the SLAMER model is an general extension from the previous model because SLAMER can handle general environment object classes.
The previous model can only consider the mapped and unmapped classes.

The most related work to SLAMER is our previous work~\cite{AkaiRA-L2020}.
In~\cite{AkaiRA-L2020}, localization using semantic information is presented.
In the method, the deep-learning-based semantic segmentation method is used as an object recognition method.
The localization method can cope with uncertainty of semantic segmentation using the class prediction model (CPM).
CPM models probabilistic distribution over the semantic segmentation results based on the off-line test.
Owing to the use of CPM, localization accuracy and robustness to inaccurate object recognition results can be improved.
However, CPM can only be used for likelihood calculation for localization, that is, spatial environmental information, such as unmeasurable walls and no entry lines which cannot be used for localization, cannot be handled.
SLAMER provides an extended framework that can be used for recognition of such spatial environmental objects.

\section{Proposed method}
\label{sec:proposed_method}

\subsection{Target problem and variable definition}
\label{subsec:target_problem_and_variable_definition}

The target problem of this paper is accurately performing localization with environment recognition results and a semantic map.
We also consider how environment recognition accuracy can be improved utilizing the semantic map.
In this work, we focus on mobile robot localization with LiDAR-based implementation.

We denote a robot pose as ${\bf x}$.
We assume that the robot is equipped with an inertial measurement system (INS) and a LiDAR.
These measurements are denoted by ${\bf u}$ and ${\bf z}$.
The semantic map is represented by a grid or voxel map, ${\bf m} = (m_{1}, ..., m_{M})$, where $M$ is number of cells and $m_{i} \in \mathcal{L}$ is a $i$th cell's environmental object label included in the environmental object list $\mathcal{L}$.

In SLAMER, an environment recognition method is used.
The recognition results and the hyperparameters of the method are denoted by $\hat{{\bf c}}$ and $\Theta$.
The recognition method can be implemented with any methods.
In this work, we assume that the recognition method outputs probability over the environmental object classes, i.e., $\hat{{\bf c}} = (\hat{{\bf c}}^{[1]}, ..., \hat{{\bf c}}^{[{}^{\rm e}K]})$, $\hat{{\bf c}}^{[k]} = (\hat{c}^{[k, 1]}, ..., \hat{c}^{[k, L]})$, $0 \leq \hat{c}^{[k, l]} \leq 1$, and $\sum_{l \in \mathcal{L}} \hat{c}^{[k, l]} = 1$, where ${}^{\rm e}K$ and $L$ are numbers of recognized environmental objects and object classes.
We further assume that the true classes, ${\bf c}$, can be estimated using the recognition results, where ${\bf c} = ({\bf c}^{[1]}, ..., {\bf c}^{[{}^{\rm e}K]})$, ${\bf c}^{[k]} = (c^{[k, 1]}, ..., c^{[k, L]})$, $c^{[k, l]} \in \{0, 1\}$, and $\sum_{l \in \mathcal{L}} c^{[k, l]} = 1$.
It should be noted that the hyperparameters depend on implementation of the recognition method and do not have an important role in this work.

\subsection{Formulation}

The graphical model of SLAMER is shown in the top of Fig.~\ref{fig:concept} (the concept of its modeling is described in Section~\ref{sec:introduction}).
The objective of SLAMER is to estimate the joint posterior distribution shown in Eq.~(\ref{eq:joint_posterior}).
\begin{align}
    p({\bf x}_{t}, {\bf c}_{t} | {\bf u}_{1:t}, {\bf z}_{1:t}, \hat{{\bf c}}_{1:t}, {\bf m}, \Theta),
    \label{eq:joint_posterior}
\end{align}
where $t$ and $1:t$ represent current and time sequence data.
Eq.~(\ref{eq:joint_posterior}) can be decomposed using the multiply theorem.
\begin{align}
    p({\bf x}_{t} | {\bf u}_{1:t}, {\bf z}_{1:t}, \hat{{\bf c}}_{1:t}, {\bf m}, \Theta)
    p({\bf c}_{t} | {\bf x}_{t}, {\bf u}_{1:t}, {\bf z}_{1:t}, \hat{{\bf c}}_{1:t}, {\bf m}, \Theta).
    \label{eq:decomposed_joint_posterior}
\end{align}
Then, we focus on how these two terms can be formulated.

The left term of Eq.~(\ref{eq:decomposed_joint_posterior}) can be re-written using the Bayes and low of total probability theorems and D-separation~\cite{Bishop:2006:PRM:1162264} as shown in Eq.~(\ref{eq:pose_distribution}).
\begin{align}
    \begin{split}
    & \eta \underbrace{ \sum_{{\bf c}_{t}} \left\{ p(\hat{{\bf c}} | {\bf c}_{t}, {\bf z}_{t}, \Theta) p({\bf c}_{t} | {\bf x}_{t}, {\bf m}) \right\} p({\bf z}_{t} | {\bf x}_{t}, {\bf m}) }_{{\rm likelihood~distribution}} \\
    & \underbrace{ \int p({\bf x}_{t} | {\bf x}_{t-1}, {\bf u}_{t}) p({\bf x}_{t-1} | {\bf u}_{1:t-1}, {\bf z}_{1:t-1}, \hat{{\bf c}}_{1:t-1}, {\bf m}, \Theta) {\rm d}{\bf x}_{t-1} }_{{\rm predictive~distribution}},
    \end{split}
    \label{eq:pose_distribution}
\end{align}
where $\eta$ is a normalization constant, $p(\hat{{\bf c}} | {\bf c}_{t}, {\bf z}_{t}, \Theta)$ is the environmental object recognition model, $p({\bf c}_{t} | {\bf x}_{t}, {\bf m})$ is the prior distribution over the environmental object classes based on the semantic map, $p({\bf z}_{t} | {\bf x}_{t}, {\bf m})$ is the measurement model, and $p({\bf x}_{t} | {\bf x}_{t-1}, {\bf u}_{t})$ is the motion model.
It should be noted that the measurement model is only used as the likelihood distribution in the general localization model presented in~\cite{Thrun:2005:PR:1121596}; however, additional two models are used in the SLAMER's likelihood distribution.
Because SLAMER contains these two distribution in the likelihood distribution, it can cope with uncertainties of environment recognition and mapping.

The right term of Eq.~(\ref{eq:decomposed_joint_posterior}) can also be re-written using the Bayes and low of total probability theorems and D-separation as shown in Eq.~(\ref{eq:object_class_distribution}).
\begin{align}
    \eta p(\hat{{\bf c}} | {\bf c}_{t}, {\bf z}_{t}, \Theta) p({\bf c}_{t} | {\bf x}_{t}, {\bf m})
    \label{eq:object_class_distribution}
\end{align}
Eq.~(\ref{eq:object_class_distribution}) shows that environment recognition based on the map is updated using the environmental object recognition model.
This update means that the true object classes are estimated based on the Bayes theorem in SLAMER.
As a result, robustness in environment recognition can also be improved.
In the next section, implementation ways are detailed.

\section{Implementation}
\label{sec:implementation}

Our target is to estimate the joint posterior distribution shown in Eq.~(\ref{eq:joint_posterior}).
To estimate the posterior, Rao-Blackwellized particle filter (RBPF) is used.
Specifically, the pose and environmental object class distributions shown in Eqs.~(\ref{eq:pose_distribution}) and (\ref{eq:object_class_distribution}) are estimated with analytical and sampling-based methods, respectively.
This estimation is achieved according to the following processes.
\begin{description}
    \item[A] update particles' pose based on the robot's motion model
    \item[B] perform environmental object recognition
    \item[C] calculate particles' likelihood
    \item[D] update environmental object recognition results with the maximum likelihood particle's pose according to Eq.~(\ref{eq:object_class_distribution})
    \item[E] estimate the pose
    \item[F] perform re-sampling if necessary and go back to A
\end{description}
This section details each process.

\subsection{Motion model}

We assume that the robot's motion model can be denoted as ${\bf x}_{t} = {\bf f}({\bf x}_{t-1}, {\bf u}_{t})$, where ${\bf f}(\cdot)$ is the motion model.
$i$th particle pose, ${\bf x}_{t}^{[i]}$, is updated as follow.
\begin{align}
    {\bf x}_{t}^{[i]} = {\bf f}({\bf x}_{t-1}^{[i]}, {\bf u}_{t}^{[i]}),~~~{\bf u}_{t}^{[i]} \sim \mathcal{N}({\bf u}_{t}, \Sigma_{\rm u}),
    \label{eq:motion_model}
\end{align}
where $\mathcal{N}({\bf u}_{t}, \Sigma_{\rm u})$ is the normal distribution with mean, ${\bf u}_{t}$, and covairnace, $\Sigma_{\rm u}$.
The particles updated with Eq.~(\ref{eq:motion_model}) approximate the predictive distribution.
In both the experiments conducted in this paper, the differential drive model is used as the motion model.

\subsection{Environmental object recognition}
\label{subsec:environmental_object_recognition}

In this work, we tested SLAMER in two cases; SemanticKITTI dataset and indoor mobile robot cases.
In these cases, we used different recognition methods.
These are detailed in Sections~\ref{subsec:deep-learning-based_environment_recognition} and \ref{subsec:object_recognition_from_2d_lidar_measurement}, respectively.
It should be noted that both the recognition methods output probability over the environment object classes, $\hat{{\bf c}}$, and it is used in the likelihood calculation and the update of the environmental object recognition results.

\subsection{Likelihood calculation}

$i$th particle's likelihood, $\omega_{t}^{[i]}$, is calculated using Eq.~(\ref{eq:slamer_likelihood}).
\begin{align}
    \begin{split}
        \omega_{t}^{[i]} \propto & \prod_{k=1}^{{}^{\rm e}K} \left( \sum_{l \in \mathcal{L}} \left\{ p(\hat{{\bf c}}_{t}^{[k]} | c_{t}^{[k, l]}, \hat{{\bf z}}_{t}^{[k]}, \Theta) p(c_{t}^{[k, l]} | {\bf x}_{t}^{[i]}, {\bf m}) \right\} \right) \\
        & \prod_{k=1}^{{}^{\rm r}K} p({\bf z}_{t}^{[k]} | {\bf x}_{t}^{[i]}, {\bf m}),
    \end{split}
    \label{eq:slamer_likelihood}
\end{align}
where ${{}^{\rm r}K}$ is the number of the LiDAR measurement used for localization.
We assumed that the environmental object recognition results and sensor measurement are independent one another, and the probabilistic models included in the likelihood distribution shown in Eq.~(\ref{eq:pose_distribution}) can be decomposed as presented in~\cite{Thrun:2005:PR:1121596}.

The prior over the object classes is modeled using the normal distribution.
\begin{align}
    p(c_{t}^{[k, l]} | {\bf x}_{t}^{[i]}, {\bf m}) = \mathcal{N}(d_{t}^{[k, l]}; 0, \sigma_{\rm d}),
    \label{eq:object_classes_prior}
\end{align}
where $d_{t}^{[k, l]}$ is a representative distance from the recognized object to the closest obstacles existing in $l$th label's semantic map.

The environmental object recognition model is modeled using the Dirichlet distribution.
\begin{align}
    \begin{split}
        & p(\hat{{\bf c}}_{t}^{[k]} | c_{t}^{[k, l]}, \hat{{\bf z}}_{t}^{[k]}, \Theta)
        = {\rm Dir}\left( \hat{{\bf c}}_{t}^{[k]}; {\bf a}(\hat{{\bf z}}_{t}^{[k]}, \Theta) \right) \\
        & = \frac{ \Gamma\left( \sum_{l \in \mathcal{L}} a^{[k, l]} \right) }{ \prod_{l \in \mathcal{L}} \Gamma(a^{[k, l]}) }
          \prod_{l \in \mathcal{L}} \left( \hat{c}_{t}^{[k, l]} \right)^{a^{[k, l]} - 1},
    \end{split}
    \label{eq:environmental_object_recognition_model}
\end{align}
where ${\bf a}(\cdot) = (a^{[k, 1]}, ..., a^{[k, L]})$, $a^{[k, l]} > 0$ is the hyperparameters and $\hat{{\bf z}}_{t}^{[k]}$ is the measurement used for recognizing $k$th environmental object.
In this work, we determined the hyperparameters as follow.
\begin{align}
    a^{[k, l]} =
    \begin{cases}
        a_{1} & {\rm if~} c_{t}^{[k, l]} = 1 \\
        a_{2} & {\rm otherwise},
    \end{cases}
    \label{eq:hyperparameters}
\end{align}
where $a_{1}$ and $a_{2}$ are arbitrary constants.
Concrete values of them are determined while respecting to performance of the recognition method.

The measurement model is modeled using the likelihood field model (LFM)~\cite{Thrun:2005:PR:1121596}.
\begin{align}
    \begin{split}
        & p({\bf z}_{t}^{[k]} | {\bf x}_{t}^{[i]}, {\bf m}) = p_{\rm lfm}({\bf z}_{t}^{[k]} | {\bf x}_{t}^{[i]}, {\bf m}) \\
        & = \left( \begin{array}{c}
            z_{\rm hit} \\
            z_{\rm max} \\
            z_{\rm rand}
        \end{array} \right)^{\top}
        \cdot
        \left( \begin{array}{c}
            p_{\rm hit}({\bf z}_{t}^{[k]} | {\bf x}_{t}^{[i]}, {\bf m}) \\
            p_{\rm max}({\bf z}_{t}^{[k]} | {\bf x}_{t}^{[i]}, {\bf m}) \\
            p_{\rm rand}({\bf z}_{t}^{[k]} | {\bf x}_{t}^{[i]}, {\bf m})
        \end{array} \right),
    \end{split}
    \label{eq:lfm}
\end{align}
where $z_{\rm hit}$, $z_{\rm max}$, and $z_{\rm rand}$ are arbitrary constants satisfying $z_{\rm hit} + z_{\rm max} + z_{\rm rand} = 1$, and $p_{\rm hit}(\cdot)$, $p_{\rm max}(\cdot)$, and $p_{\rm rand}(\cdot)$ are denoted as follow.
\begin{gather}
    p_{\rm hit}({\bf z}_{t}^{[k]} | {\bf x}_{t}^{[i]}, {\bf m}) = \mathcal{N} \left(d({\bf x}_{t}^{[i]}, {\bf z}_{t}^{[k]}, {\bf m}); 0, \sigma^{2} \right), \\
    p_{\rm max}({\bf z}_{t}^{[k]} | {\bf x}_{t}^{[i]}, {\bf m}) = 
    \begin{cases} 
        1 & {\rm if~} r_{t}^{[k]} = r_{\rm max} \\
        0 & {\rm otherwise},
    \end{cases} \\
    p_{\rm rand}({\bf z}_{t}^{[k]} | {\bf x}_{t}^{[i]}, {\bf m}) = {\rm unif}(0, r_{\rm max}),
\end{gather}
where $d({\bf x}_{t}^{[i]}, {\bf z}_{t}^{[k]}, {\bf m})$ is a function that returns a distance from $k$th measurement point to the closest obstacle existing on the map, $\sigma$ is the variance, $r_{t}^{[k]}$ and $r_{\rm max}$ are $k$th and the maximum measurement ranges, and ${\rm unif}(\cdot)$ is a uniform distribution within a given range.
It should be noted that the semantics is not considered in calculation of LFM because the SLAMER model does not assume that the environmental object recognition results do not depend on the semantic map.
Other measurement models such as the beam model~\cite{Thrun:2005:PR:1121596} can be used for implementing the measurement model, but we use LFM because of its efficiency in this work.

\subsection{Update environmental object recognition results}

Environmental object recognition is performed based on the method described in Section~\ref{subsec:environmental_object_recognition}.
Then, Eqs.~(\ref{eq:object_classes_prior}) and (\ref{eq:environmental_object_recognition_model}) are used to obtain the posterior over the true object classes.
It should be noted that the pose ${\bf x}_{t}^{[i]}$ is replaced to that of the maximum likelihood particle in this update and this updated result is used as the final environment recognition results.

\subsection{Pose estimate}

The particles' likelihood is normalized.
Then, the weighted average regarding the pose is calculated and it is used as the estimated pose.

\subsection{Re-sampling}

The effective sample size, $1 / \sum_{i=1}^{M} {\left( \omega_{t}^{[i]} \right)^{2}}$, is first calculated, where $M$ is the number of the particles.
If it is less than $M / 2$, re-sampling is performed.

\section{Experiments with dataset}
\label{sec:dataset_experiments}

In this section, we evaluate SLAMER of 3D LiDAR-based implementation on the SemanticKITTI dataset~\cite{behley2019iccv}.

\subsection{Setup}

The SemanticKITTI dataset has vehicle trajectories and corresponding 3D point clouds.
The ground truth object labels are assigned to each scan point.
We first plotted the 3D point clouds obtained from the static objects such as buildings and roads according to the vehicle trajectory and built 3D maps.
The method presented in~\cite{AkaiIV2020} was used for building a distance field map which enables to efficiently get the closest distance from the obstacles.
We then simulated noisy odometry measurements from the trajectories.
In the experiment step, the noisy odometry was given to update the vehicle pose and accumulated errors were compensated by matching of the 3D point clouds with the map.
The estimated poses were compared with the given trajectories and estimation accuracy was calculated.

500 particles were used to implement RBPF.
The hyperparameters shown in Eq.~(\ref{eq:hyperparameters}) were experimentally set to $a_{1} = 1.2$ and $a_{2} = 1$.

\subsection{Deep-learning-based environment recognition}
\label{subsec:deep-learning-based_environment_recognition}

We used the same environment recognition method presented in~\cite{AkaiRA-L2020}.
In~\cite{AkaiRA-L2020}, the SegNet~\cite{badrinarayanan2015segnet}-based method was used.
The depth and intensity maps are made from the 3D point cloud and these maps are fed to the network.
The network infers probability over the object classes for each pixel and the output probabilities are treated as $\hat{{\bf c}}$.
The probability is calculated via the softmax function.

\subsection{Comparison methods for pose estimation}

\subsubsection{Likelihood field model (LFM)}

The general localization model presented in~\cite{Thrun:2005:PR:1121596} is formulated as the recursive Bayes filter as shown in Eq.~(\ref{eq:general_localization_model}).
\begin{align}
    \begin{split}
        & p({\bf x}_{t} | {\bf u}_{1:t}, {\bf z}_{1:t}, {\bf m}) = \eta p({\bf z}_{t} | {\bf x}_{t}, {\bf m}) \\
        & \int p({\bf x}_{t} | {\bf x}_{t-1}, {\bf u}_{t}) p({\bf x}_{t-1} | {\bf u}_{1:t-1}, {\bf z}_{1:t-1}, {\bf m}) {\rm d}{\bf x}_{t-1}.
    \end{split}
    \label{eq:general_localization_model}
\end{align}
We implemented LFM~\cite{Thrun:2005:PR:1121596} as the measurement model to estimate Eq.~(\ref{eq:general_localization_model}).
\begin{align}
    p({\bf z}_{t} | {\bf x}_{t}, {\bf m}) = \prod_{k=1}^{K} p_{\rm lfm}({\bf z}_{t}^{[k]} | {\bf x}_{t}, {\bf m})
    \label{eq:decomposed_lfm}
\end{align}

\subsubsection{Semantic likelihood field model (SLFM)}

We also implemented LFM with the estimated object classes by the network to estimate Eq.~(\ref{eq:general_localization_model}).
\begin{align}
    p({\bf z}_{t} | {\bf x}_{t}, {\bf m}) = \prod_{k=1}^{K} \prod_{l \in \mathcal{L}} p_{\rm slfm}({\bf z}_{t}^{[k]} | {\bf x}_{t}, {\bf m}_{l})^{\mathds{1} \left( {\rm if}~\hat{c}^{[k, l]}~{\rm is~max~in}~\hat{\bf c}^{[k]} \right)},
\end{align}
where $\mathds{1}(\cdot)$ is an indicator function which is equal to 1 when the condition within the bracket is true, and 0 otherwise, and $p_{\rm slfm}({\bf z}_{t}^{[k]} | {\bf x}_{t}, {\bf m})$ is denoted as follow.
\begin{align}
    p_{\rm slfm}({\bf z}_{t}^{[k]} | {\bf x}_{t}, {\bf m}_{l}) =
    \begin{cases}
        \frac{\lambda \exp\left(-\lambda r_{t}^{[k]} \right)}{1 - \exp\left(-\lambda r_{\rm max}\right)} & {\rm if}~l \in {\rm unknown} \\
        p_{\rm lfm}({\bf z}_{t}^{[k]} | {\bf x}_{t}, {\bf m}_{l}) & {\rm otherwise},
    \end{cases}
\end{align}
where $\lambda$ is the hyperparameter for the exponential distribution, and ${\bf m}_{l}$ is $l$th label's semantic map.
We refer this comparison method to the semantic likelihood field model (SLFM).

\subsubsection{Class prediction model (CPM)}

The class prediction model (CPM), $p(\hat{{\bf c}}_{t} | {\bf x}_{t}, {\bf z}_{t}, {\bf m}, \Theta)$, is presented in~\cite{AkaiRA-L2020} and it is used for estimating the posterior over the pose shown in Eq.~(\ref{eq:cpm_localization_model}).
\begin{align}
    \begin{split}
        & p({\bf x}_{t} | {\bf u}_{1:t}, {\bf z}_{1:t}, {\bf m}, \Theta) = \eta p(\hat{{\bf c}}_{t} | {\bf x}_{t}, {\bf z}_{t}, {\bf m}, \Theta) \\
        & \int p({\bf x}_{t} | {\bf x}_{t-1}, {\bf u}_{t}) p({\bf x}_{t-1} | {\bf u}_{1:t-1}, {\bf z}_{1:t-1}, \hat{{\bf c}}_{1:t-1} {\bf m}, \Theta) {\rm d}{\bf x}_{t-1}.
    \end{split}
    \label{eq:cpm_localization_model}
\end{align}
CPM is also decomposed to ensure feasibility of calculation.
Implementation details of CPM can be seen at~\cite{AkaiRA-L2020}.

\subsection{Comparison method for environment recognition}

We compared SLAMER's environment recognition accuracy with a simple map-based recognition method.
The simple method uses SLFM.
SLFM is calculated according to the estimated pose by SLAMER.
The maximum likelihood class is assigned as the estimated class to each scan point.

\subsection{Results}

Table~\ref{tab:comparison_results_kitti} shows the comparison results.
The performance of LFM and SLFM were not similar, but it was difficult to say which one is superior.
This result indicated that the simple use of the environment recognition results cannot yield performance improvement.
However, the performance could be improved if an accurate recognition method can be available.

SLAMER outperformed LFM and SLFM in all the sequences because SLAMER can utilize the environment recognition results even though SLFM cannot.
This result showed that SLAMER can cope with uncertainty in the environment recognition.

However, SLAMER was not superior to CPM in all the sequences in terms of the localization accuracy.
This result is natural if we consider the background of their modeling.
SLAMER has an advantage besides localization performance improvement.
We discuss regarding them in Section~\ref{subsec:discussion}.

\begin{table*}[!t]
\begin{center}
\caption{Pose/angle estimation errors by LFM, SLFM, CPM, and SLAMER on the SemanticKITTI dataset. The units of the pose and angle estimation errors are centimeters and degrees.}
\begin{tabular}{|c|c|c|c|c|c|c|c|c|}
\hline
                                                           & Sequence                                                      & 03                                                                                 & 04                                                                                 & 05                                                                               & 06                                                                                   & 07                                                                                  & 09                                                                                   & 10                                                                               \\ \hline
LFM                                                        & \begin{tabular}[c]{@{}c@{}}Ave\\ Std\\ Max\end{tabular}       & \begin{tabular}[c]{@{}c@{}}15.44 / 0.38\\ 14.58 / 0.40\\ 89.10 / 3.32\end{tabular} & \begin{tabular}[c]{@{}c@{}}12.44 / 0.35\\ 11.91 / 0.30\\ 61.16 / 1.90\end{tabular} & \begin{tabular}[c]{@{}c@{}}7.88 / 0.22\\ 7.23 / 0.23\\ 98.15 / 2.40\end{tabular} & \begin{tabular}[c]{@{}c@{}}25.32 / 0.53\\ 46.57 / 1.29\\ 278.01 / 15.11\end{tabular} & \begin{tabular}[c]{@{}c@{}}29.28 / 0.44\\ 22.89 / 0.39\\ 140.36 / 2.34\end{tabular} & \begin{tabular}[c]{@{}c@{}}18.42 / 0.35\\ 55.81 / 0.74\\ 587.30 / 11.21\end{tabular} & \begin{tabular}[c]{@{}c@{}}6.16 / 0.20\\ 4.60 / 0.22\\ 48.31 / 2.34\end{tabular} \\ \hline
SLFM                                                       & \begin{tabular}[c]{@{}c@{}}Ave\\ Std\\ Max\end{tabular}       & \begin{tabular}[c]{@{}c@{}}11.47 / 0.32\\ 7.79 / 0.29\\ 54.18 / 1.88\end{tabular}  & \begin{tabular}[c]{@{}c@{}}14.75 / 0.26\\ 12.24 / 0.24\\ 77.01 / 1.88\end{tabular} & \begin{tabular}[c]{@{}c@{}}9.28 / 0.26\\ 6.79 / 0.25\\ 65.46 / 2.44\end{tabular} & \begin{tabular}[c]{@{}c@{}}17.88 / 0.42\\ 13.56 / 0.40\\ 100.39 / 2.69\end{tabular}  & \begin{tabular}[c]{@{}c@{}}11.23 / 0.33\\ 7.52 / 0.29\\ 64.94 / 2.20\end{tabular}   & \begin{tabular}[c]{@{}c@{}}14.06 / 0.39\\ 10.86 / 0.35\\ 104.53 / 3.32\end{tabular}  & \begin{tabular}[c]{@{}c@{}}9.81 / 0.31\\ 6.99 / 0.28\\ 86.76 / 2.29\end{tabular} \\ \hline
CPM                                                        & \begin{tabular}[c]{@{}c@{}}Ave\\ Std\\ Max\end{tabular}       & \begin{tabular}[c]{@{}c@{}}7.80 / 0.20\\ 5.10 / 0.18\\ 38.36 / 1.61\end{tabular}   & \begin{tabular}[c]{@{}c@{}}7.52 / 0.14\\ 5.57 / 0.12\\ 39.79 / 0.78\end{tabular}   & \begin{tabular}[c]{@{}c@{}}5.42 / 0.14\\ 3.32 / 0.15\\ 25.45 / 1.29\end{tabular} & \begin{tabular}[c]{@{}c@{}}9.58 / 0.22\\ 6.71 / 0.20\\ 58.98 / 1.39\end{tabular}     & \begin{tabular}[c]{@{}c@{}}6.75 / 0.19\\ 5.37 / 0.19\\ 69.14 / 1.62\end{tabular}    & \begin{tabular}[c]{@{}c@{}}7.75 / 0.19\\ 5.09 / 0.17\\ 46.84 / 1.31\end{tabular}     & \begin{tabular}[c]{@{}c@{}}5.65 / 0.17\\ 3.85 / 0.16\\ 41.25 / 1.27\end{tabular} \\ \hline
SLAMER                                                     & \begin{tabular}[c]{@{}c@{}}Ave\\ Std\\ Max\end{tabular}       & \begin{tabular}[c]{@{}c@{}}9.97 / 0.25\\ 7.34 / 0.24\\ 68.82 / 2.54\end{tabular}   & \begin{tabular}[c]{@{}c@{}}8.69 / 0.15\\ 6.29 / 0.13\\ 43.52 / 0.70\end{tabular}   & \begin{tabular}[c]{@{}c@{}}7.60 / 0.20\\ 5.13 / 0.18\\ 37.21 / 1.37\end{tabular} & \begin{tabular}[c]{@{}c@{}}13.41 / 0.28\\ 8.76 / 0.25\\ 65.33 / 1.69\end{tabular}    & \begin{tabular}[c]{@{}c@{}}10.51 / 0.28\\ 6.67 / 0.24\\ 45.42 / 1.72\end{tabular}   & \begin{tabular}[c]{@{}c@{}}12.12 / 0.31\\ 9.49 / 0.27\\ 131.92 / 2.50\end{tabular}   & \begin{tabular}[c]{@{}c@{}}7.85 / 0.24\\ 6.06 / 0.23\\ 43.82 / 2.05\end{tabular} \\ \hline \hline
ER Acc                                                     & \begin{tabular}[c]{@{}c@{}}Ave\\ Std\\ Min\\ Max\end{tabular} & \begin{tabular}[c]{@{}c@{}}76.90 \%\\ 6.50 \%\\ 60.33 \%\\ 92.09 \%\end{tabular}   & \begin{tabular}[c]{@{}c@{}}77.35 \%\\ 5.65 \%\\ 60.17 \%\\ 91.01 \%\end{tabular}   & \begin{tabular}[c]{@{}c@{}}79.04 \%\\ 4.07 \%\\ 63.49 \%\\ 89.84 \%\end{tabular} & \begin{tabular}[c]{@{}c@{}}74.32 \%\\ 6.76 \%\\ 54.71 \%\\ 86.16 \%\end{tabular}     & \begin{tabular}[c]{@{}c@{}}85.30 \%\\ 4.67 \%\\ 70.51 \%\\ 95.07 \%\end{tabular}    & \begin{tabular}[c]{@{}c@{}}78.61 \%\\ 5.30 \%\\ 57.36 \%\\ 90.41 \%\end{tabular}     & \begin{tabular}[c]{@{}c@{}}75.21 \%\\ 5.80 \%\\ 44.80 \%\\ 85.99 \%\end{tabular} \\ \hline
\begin{tabular}[c]{@{}c@{}}Map-based\\ ER Acc\end{tabular} & \begin{tabular}[c]{@{}c@{}}Ave\\ Std\\ Min\\ Max\end{tabular} & \begin{tabular}[c]{@{}c@{}}83.52 \%\\ 6.21 \%\\ 70.20 \%\\ 95.69 \%\end{tabular}   & \begin{tabular}[c]{@{}c@{}}87.34 \%\\ 6.34 \%\\ 74.21 \%\\ 96.77 \%\end{tabular}   & \begin{tabular}[c]{@{}c@{}}78.25 \%\\ 7.94 \%\\ 46.90 \%\\ 94.48 \%\end{tabular} & \begin{tabular}[c]{@{}c@{}}75.59 \%\\ 10.36 \%\\ 43.25 \%\\ 92.84 \%\end{tabular}    & \begin{tabular}[c]{@{}c@{}}79.75 \%\\ 8.61 \%\\ 44.95 \%\\ 93.35 \%\end{tabular}    & \begin{tabular}[c]{@{}c@{}}77.59 \%\\ 7.52 \%\\ 52.29 \%\\ 95.27 \%\end{tabular}     & \begin{tabular}[c]{@{}c@{}}76.15 \%\\ 8.61 \%\\ 40.34 \%\\ 94.76 \%\end{tabular} \\ \hline
\begin{tabular}[c]{@{}c@{}}SLAMER\\ ER Acc\end{tabular}    & \begin{tabular}[c]{@{}c@{}}Ave\\ Std\\ Min\\ Max\end{tabular} & \begin{tabular}[c]{@{}c@{}}85.72 \%\\ 6.42 \%\\ 70.22 \%\\ 95.63 \%\end{tabular}   & \begin{tabular}[c]{@{}c@{}}90.15 \%\\ 5.28 \%\\ 75.63 \%\\ 96.64 \%\end{tabular}   & \begin{tabular}[c]{@{}c@{}}83.86 \%\\ 7.79 \%\\ 49.28 \%\\ 95.69 \%\end{tabular} & \begin{tabular}[c]{@{}c@{}}81.17 \%\\ 8.80 \%\\ 47.41 \%\\ 94.31 \%\end{tabular}     & \begin{tabular}[c]{@{}c@{}}89.02 \%\\ 2.95 \%\\ 76.51 \%\\ 95.82 \%\end{tabular}    & \begin{tabular}[c]{@{}c@{}}80.32 \%\\ 7.80 \%\\ 53.38 \%\\ 95.82 \%\end{tabular}     & \begin{tabular}[c]{@{}c@{}}79.21 \%\\ 6.77 \%\\ 54.41 \%\\ 94.68 \%\end{tabular} \\ \hline
\end{tabular}
\label{tab:comparison_results_kitti}
\end{center}
\end{table*}

Table~\ref{tab:comparison_results_kitti} also shows the environment recognition accuracy by the network (ER Acc), simple method (Map-based ER ACC), and SLAMER (SLAMER ER Acc), respectively.
SLAMER achieved accurate recognition more than the network in all the sequences even when the simple map-based method sometimes degradated.
However, the minimum recognition accuracy by SLAMER was sometimes bad more than that of the network.
This was yielded by an inaccurate localization result.
However, SLAMER could improved the average recognition accuracy.
In addition, we could confirm that the simple-based environment recognition is not effective to improve the recognition accuracy.
From these results, we could reveal that SLAMER can cope with environment recognition and mapping uncertainties.

\subsection{Discussion}
\label{subsec:discussion}

As can be seen from Table~\ref{tab:comparison_results_kitti}, CPM outperformed all the methods in terms of the localization accuracy.
We discuss why SLAMER cannot outperform CPM while respecting the advantage of SLAMER.

The model that derives CPM assumes that the environment recognition results, $\hat{{\bf c}}$, depend on the pose, sensor measurement, and map.
This dependency enables to consider how the environmental object classes are predicted while considering relationship between the pose, sensor measurement, and map.
This relationship is significant for localization.
Hence, CPM, $p(\hat{{\bf c}}_{t} | {\bf x}_{t}, {\bf z}_{t}, {\bf m}, \Theta)$, can utilize the class prediction results to improve the localization accuracy.

However, the likelihood distribution used in SLAMER, $\sum_{{\bf c}_{t}} \left\{ p(\hat{{\bf c}} | {\bf c}_{t}, {\bf z}_{t}, \Theta) p({\bf c}_{t} | {\bf x}_{t}, {\bf m}) \right\} p({\bf z}_{t} | {\bf x}_{t}, {\bf m})$, does not consider such relationship.
The environment recognition model, $p(\hat{{\bf c}} | {\bf c}_{t}, {\bf z}_{t}, \Theta)$, only considers the sensor measurement and the true classes.
Consequently, the likelihood distribution cannot have significant effect to improve the localization accuracy.
However, the likelihood distribution includes the prior distribution over the true object classes.
This contributes to improve the localization performance while utilizing the environment recognition results.

In addition, since the likelihood distribution does not have the relationship considered in CPM, objects which might not have better influence for localization can also be treated in the SLAMER framework.
For example spatial objects such as no entty lines can be handled in the framework.
This advantage is shown in the next section.

\section{Indoor experiments}
\label{sec:indoor_experiments}

In this section, we show qualitative performance of 2D LiDAR-based SLAMER with our experimental platform.
This 2D LiDAR-based implementation is publicly available\footnote{\url{https://github.com/NaokiAkai/als_ros}}.

\subsection{Experimental equipment}

\begin{figure}[!t]
    \begin{center}
        \includegraphics[width = 70 mm]{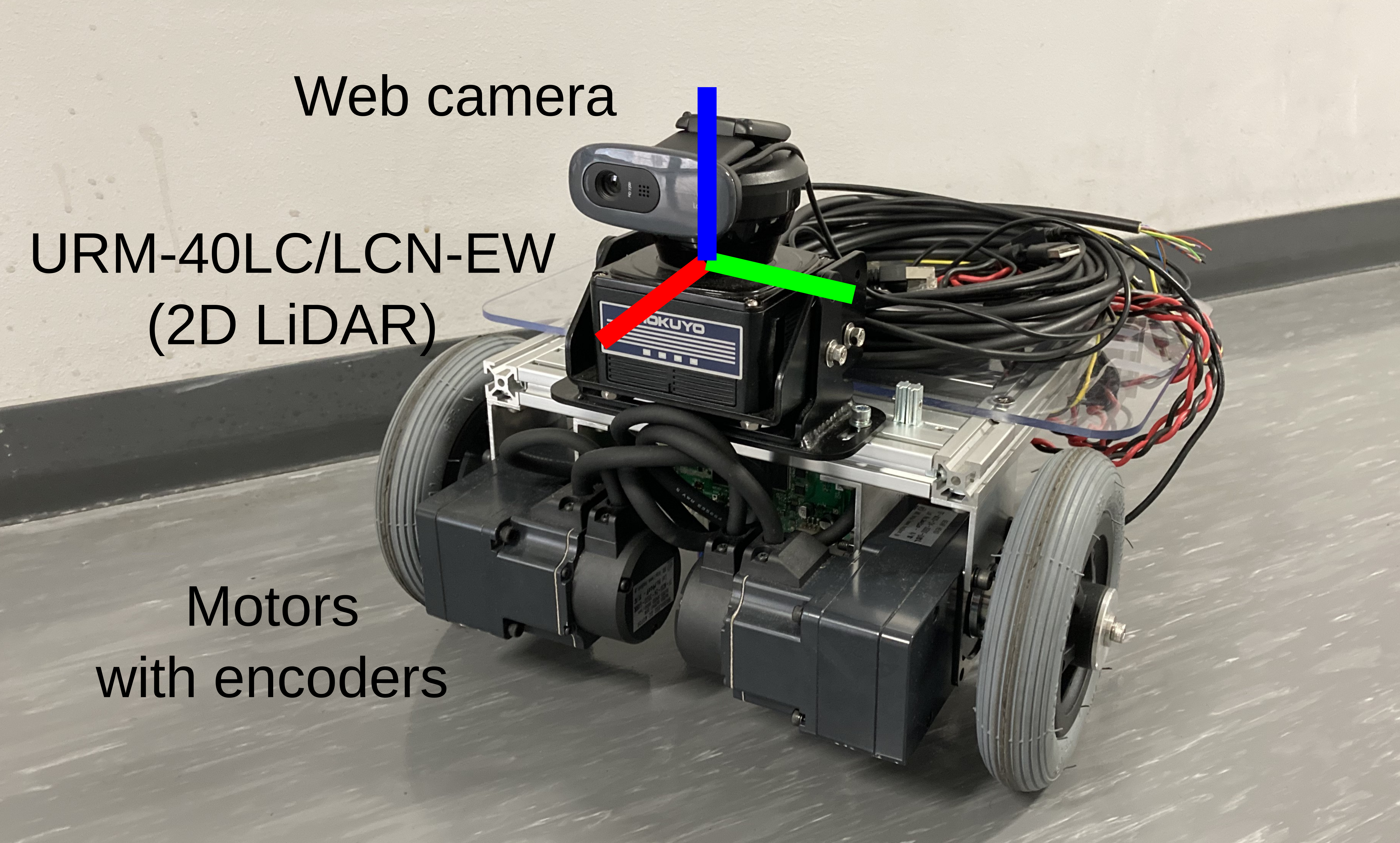}
        \caption{The experimental platform (iCart-mini). The red, green, and blue lines indicate $x$, $y$, and $z$ axes of the 2D LiDAR. The web camera is just used for visualization as shown in Fig.~\ref{fig:concept}.}
        \label{fig:icart_mini}
    \end{center}
\end{figure}

\begin{figure}[!t]
    \begin{center}
        \includegraphics[width = 80 mm]{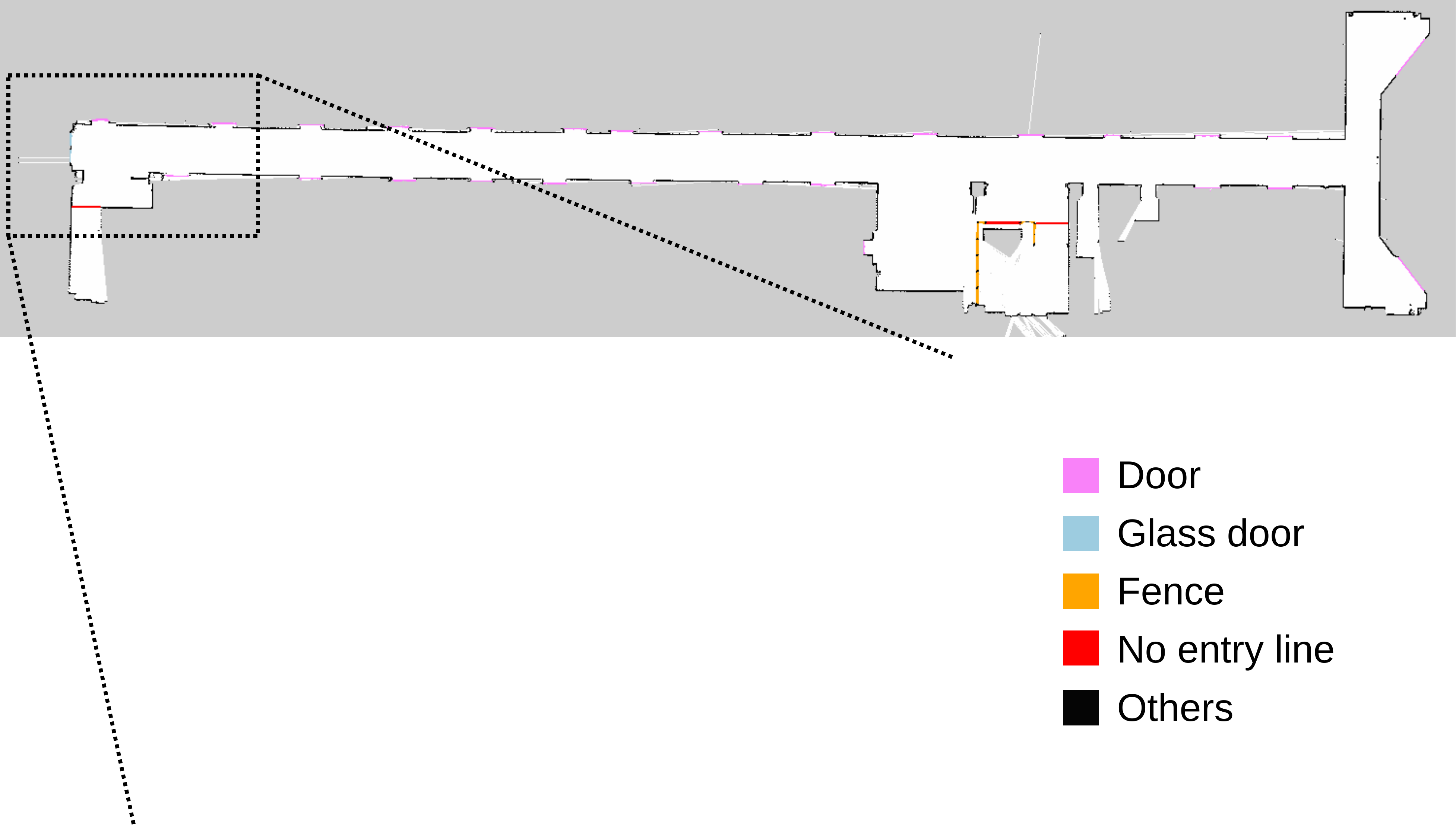}
        \caption{An example of the semantic map. The map shown in Fig.~\ref{fig:concept} is enlarged at the bottom.}
        \label{fig:semantic_map}
    \end{center}
\end{figure}

The experimental platform is shown in Fig.~\ref{fig:icart_mini}.
We used i-Cart mini robot\footnote{\url{http://wiki.ros.org/icart_mini}} equipped with URM-40LC/LCN-EW LiDAR\footnote{\url{https://www.hokuyo-aut.co.jp/search/single.php?serial=189}}.
This robot is equipped with wheel encoders and we used it as INS.
We used gmapping\footnote{\url{http://wiki.ros.org/gmapping}} to build 2D map and manually created the semantic map.
Figure~\ref{fig:semantic_map} shows an example of the semantic map.
In the experiment, door (pink), glass door (cyan), fence (orange), and no entry line (red) objects are assigned to the map.
Based on the objects, we consider following environmental object classes; open door, close door, open glass door, close glass door, no entry line, fence, free space, and others.

200 particles were used to implement RBPF.
The hyperparameters shown in Eq.~(\ref{eq:hyperparameters}) were experimentally set to $a_{1} = 2$ and $a_{2} = 2$.

\subsection{Object recognition from 2D LiDAR measurement}
\label{subsec:object_recognition_from_2d_lidar_measurement}

In this experiment, we use a simple line object detection method.
A 2D LiDAR scan image is first made and the probabilistic Hough transform implemented in OpenCV is applied to the image to detect line objects.
Then, rate of the scan points included in each line is calculated.
Probability over the environmental object classes is calculated based on the rate.
Of course, this rate-based probability calculation is inaccurate and it does not have an important role in this experiment.
Hence, we implemented a simple rule-based classification based on the rate.

We also consider spatial objects that do not have physical shape but has environmental meanings such as open doors and no entry lines.
To detect candidates of the spatial objects, we first build histogram regarding incline of the detected lines by the Hough transform.
Then, we create lines between all the scan points and these lines are referred to spatial line objects.
Incline of the spatial line objects is computed and it is compared with the incline histogram.
If the probability of the target incline is less than threshold set to the histogram, the target spatial object line is deleted.
The spatial lines which are not passed through by the scan lines are also deleted since there might be no spaces behind the lines.
Finally, the points rate is also calculated to the spatial object lines and classification is performed.
This classification is also implemented based on a simple rule.

\begin{figure}[!t]
    \begin{center}
        \includegraphics[width = 75 mm]{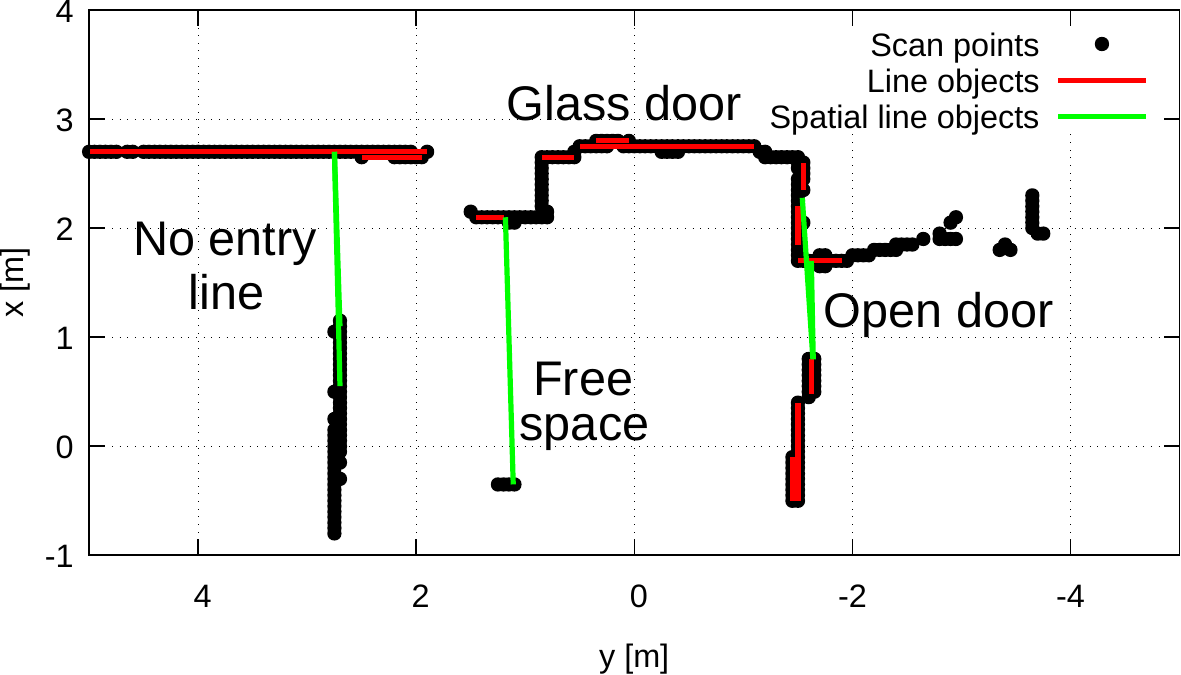}
        \caption{An example of the line object detection result in the 2D LiDAR coordinates. This example is extracted from the data shown in Fig.~\ref{fig:concept}.}
        \label{fig:line_detection_result}
    \end{center}
\end{figure}

Figure~\ref{fig:line_detection_result} shows an example of the line object detection result.
This example is extracted from the data shown in the bottom of Fig.~\ref{fig:concept}.
The black points depict the scan points and the red and green lines depict the physical and spatial line objects.
Parallel lines to the scan points were detected as the spatial line objects.

\subsection{Results and discussion}

The bottom of Fig.~\ref{fig:concept} shows the qualitative results of SLAMER's estimate.
As can be seen from the figure, we could confirm that SLAMER can recognize the environmental objects while using the map assist.
However, as we mentioned in the previous subsection, the classification based on the points rate was inaccurate.
This means that the map-based recognition was dominant in the cases shown in Fig.~\ref{fig:concept} since localization has succeeded.
We discuss SLAMER's performance based on inaccurate localization cases because performance regarding uncertainty consideration can be confirmed in such cases.

\begin{figure}[!t]
    \begin{center}
        \includegraphics[width = 85 mm]{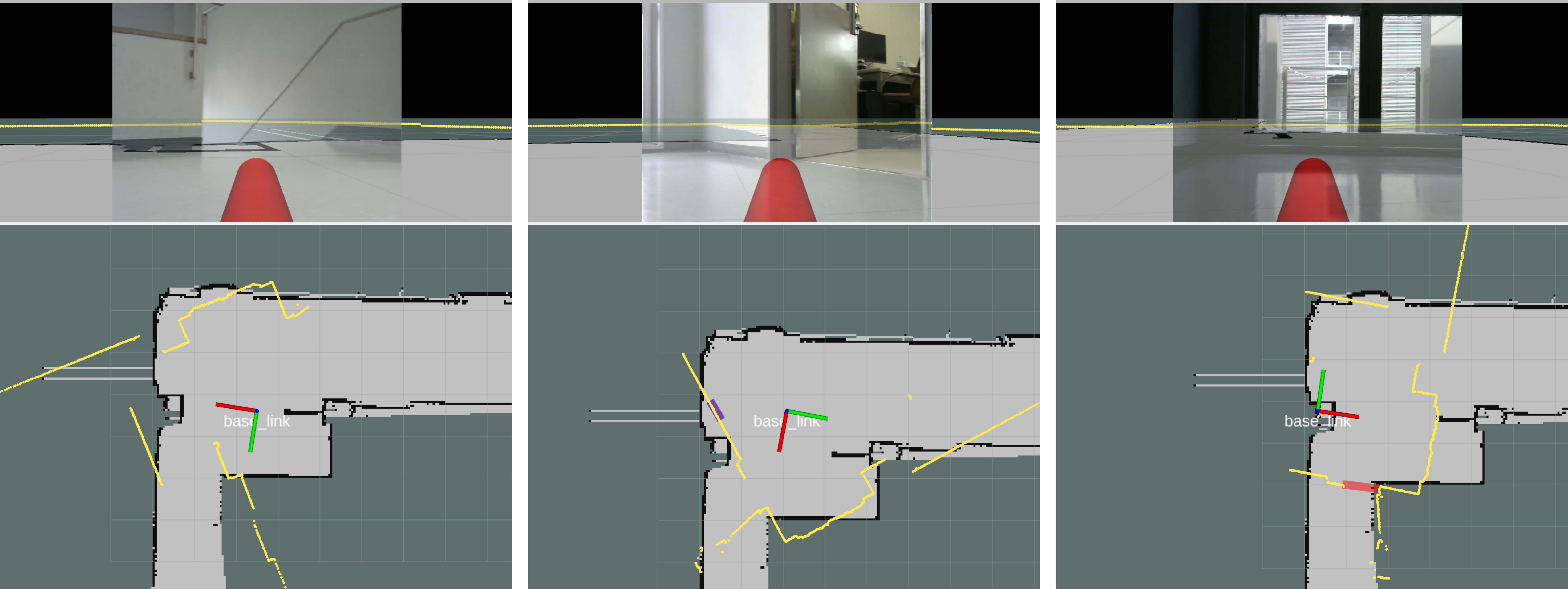}
        \caption{Environment object recognition results in miss localization cases.}
        \label{fig:miss_localization_cases}
    \end{center}
\end{figure}

Figure~\ref{fig:miss_localization_cases} shows the environmental object recognition results in miss localization cases.
The left and middle figures of Fig.~\ref{fig:miss_localization_cases} show better results of SLAMER's estimate.
The object recognition method presented in Section~\ref{subsec:object_recognition_from_2d_lidar_measurement} recognized line objects as shown in Fig.~\ref{fig:line_detection_result}; however, SLAMER did not output any environmental objects.
In particular, the open door was not recognized as the no entry line even though its position on the map was closed to the no entry line.
This result revealed that SLAMER can cope with uncertainties in localization, environment recognition, and mapping.

However, SLAMER does not work in some cases.
The right figure of Fig.~\ref{fig:miss_localization_cases} shows an miss recognition result.
The position of the open door was exactly overlapped with that of the no entry line and the open door was recognized as the no entry line.
Even though SLAMER can perform environmental object recognition with the Bayes filter, SLAMER cannot work in such a worst case.
However, occuring such a worst overlap is seldom rare. 
In addition, SLAMER can have possibility to overcome such cases because it can use the environmental object recognition model to update the map-based prior.

\section{Conclusion}
\label{sec:conclusion}

This paper has presented SLAMER, the simultaneous localization and map-assisted object recognition method.
SLAMER is the probabilistic model to cope with uncertainties included in localization, environmental object recognition, and mapping.
In this paper, we demonstrated 2D- and 3D-LiDAR-based implementation of SLAMER.
For the 3D-LiDAR-based demonstration, we used the SemanticKITTI dataset and showed that SLAMER improved both localization and environmental object recognition accuracy from that of the general methods.
For the 2D-LiDAR-based demonstration, we used the indoor mobile robot and showed that SLAMER realized recognition of unmeasurable environmental objects such as open doors and no entry lines.

We also showed that SLAMER cannot outperform the class prediction model (CPM) presented in~\cite{AkaiRA-L2020} in terms of localization accuracy because CPM assumes strong relationship between the environmental object recognition results and the semantic map.
SLAMER can have an advantage that objects which might not have significant influence to localization can be handled owing to the lack of the strong relationship.
Outperforming CPM while ensuring the advantage of SLAMER is our future work.

\section*{ACKNOWLEDGMENT}

Support for this work was given by the Toyota Motor Corporation (TMC) and JSPS KAKENHI under Grant 18K13727.
However, note that this article solely reflects the opinions and conclusions of its author and not TMC or any other Toyota entity.

\bibliographystyle{unsrt}
\bibliography{reference.bib}


\end{document}